\documentclass[journal abbreviation, manuscript]{copernicus}

\usepackage{subcaption}
\usepackage{hyperref}
\AtBeginDocument{\nolinenumbers}
\begin{document}
\bibliographystyle{copernicus}
\title{Democracy of AI numerical weather model: An example of  running the global forecasting using FourCastNet and GPUs}


\Author[1]{Iman}{Khadir} 
\Author[1]{Shane}{Stevenson}
\Author[2]{Henry}{Li}
\Author[2]{Kyle}{Krick}
\Author[1]{Abram}{Burrows}
\Author[3]{David}{Hall}
\Author[3]{Stan}{Posey}
\Author[1][sshen@sdsu.edu]{Samuel}{S. P. Shen}

\affil[1]{Department of Mathematics and Statistics,  San Diego State University, San Diego, CA 92182, USA}
\affil[2]{Research \& Cyberinfrastructure Division, San Diego State University, San Diego, CA 92182, USA}
\affil[3]{NVIDIA Corporation, Santa Clara, CA 95051, USA}




\runningtitle{TEXT}

\runningauthor{TEXT}

\received{}
\pubdiscuss{} 
\revised{}
\accepted{}
\published{}


\firstpage{1}

\maketitle

\begin{abstract}
This paper demonstrates the feasibility of democratizing AI-driven global weather forecasting models among university research groups by leveraging Graphics Processing Units (GPUs) and freely available AI models, such as NVIDIA's FourCastNetv2.
FourCastNetv2 is an NVIDIA's advanced neural network for weather prediction and is trained on a 73-channel subset of the European Centre for Medium-Range Weather Forecasts (ECMWF) Reanalysis v5 (ERA5) dataset at single levels and different pressure levels. Although the training specifications for FourCastNetv2 are not released to the public, the training documentation of the model's first generation, FourCastNet, is available to all users. The training had 64 A100 GPUs and took 16 hours to complete. Although NVIDIA's models offer significant reductions in both time and cost compared to traditional Numerical Weather Prediction (NWP), reproducing published forecasting results presents ongoing challenges for resource-constrained university research groups with limited GPU availability. We demonstrate both (i) leveraging FourCastNetv2 to create predictions through the designated application programming interface (API), and (ii) utilizing NVIDIA hardware to train the original FourCastNet model. Further, this paper demonstrates the capabilities and limitations of NVIDIA A100's for resource-limited research groups in universities. We also explore data management, training efficiency, and model validation, highlighting the advantages and challenges of using limited high-performance computing resources. Consequently, this paper and its corresponding GitHub materials may serve as an initial guide for  other university research groups and courses related to machine learning, climate science, and data science to develop research and education programs on AI weather forecasting, and hence help democratize AI NWP in the digital economy.
\end{abstract}


\introduction  
High-resolution global weather forecasting traditionally demands supercomputing-level resources due to the extensive computations involved in the solution of complex mathematical models and the analysis of vast datasets.
Hence, the enterprises of numerical weather forecasting have been limited to major governmental agencies and research centers. It is infeasible for small university research labs to make operational, high-resolution weather forecasts.

The rapid and notable advancement of AI in weather forecasting has dramatically changed the situation. Modern GPUs, such as NVIDIA's A100 and H200 are affordable by universities and can act as supercomputers on which AI algorithms can run by university research groups. The trend of this remarkable transition of high-resolution numerical weather forecasts from major computing centers to university research labs, or even to university classrooms, has provided unprecedented opportunities for higher education. In particular, both graduate and undergraduate students can practice operational numerical forecasting in their research or coursework. However, a gap exists between the accessibility and usability of advanced AI weather forecasting tools like NVIDIA's FourCastNet \citep{pathak2022fourcastnetglobaldatadrivenhighresolution} and DeepMind's GenCast \citep{lam2023graphcastlearningskillfulmediumrange} for university researchers with limited resources and/or prior AI expertise.
The purpose of this paper is to fill this gap by providing users' guidance for university research labs to run or even to train AI weather forecasting models. We are writing this paper using NVIDIA's FourCastNetv2 as an example.

NVIDIA is a leader in HPC system development and software technologies that support cutting-edge AI weather forecasting. The initial release of FourCastNet \citep{pathak2022fourcastnetglobaldatadrivenhighresolution} was the first major model of this kind to be published.
This is a data-driven forecast that uses an Adaptive Neural Fourier Operator \citep{guibas2022adaptivefourierneuraloperators}. Many other AI weather forecasting models have appeared since then, such as Google's GraphCast \citep{lam2023graphcastlearningskillfulmediumrange}, GenCast \citep{price2024gencastdiffusionbasedensembleforecasting}, Microsoft's Aurora \citep{bodnar2024aurora}, and Huawei's Pangu-Weather \citep{bi2023accurate}. Each model has its own specific features and merits. In 2023, NVIDIA released FourCastNetv2, which has a distinct feature of using a Spherical Fourier Neural Operator \citep{bonev2023sphericalfourierneuraloperators}. This updated Version-2 of FourCastNet (FourCastNetv2) improved model stability and accuracy. FourCastNet is part of NVIDIA's efforts to create a digital twin of the Earth through an accessible platform called Earth-2. Earth-2 weather analytics brings together the latest HPC climate technologies, which will allow for scientists, governments, and enthusiasts to better prepare for weather and climate related disasters \citep{earth2}. An example of the visualization tools Earth-2 Weather Analytics provides is a graphical user interface, where users can generate predictions using FourCastNetv2 and view them on an interactive, 3D globe that can be rotated, zoomed, and explored from any angle \citep{build_nvidia}.

Our paper is written based upon FourCastNetv2 for the purpose of tutoring university research groups to make more AI weather forecasts. It is not our intention to evaluate the contributions of each of these AI forecasting models. Instead, we are providing (i) a step-by-step guide to other groups and universities to train students to make AI weather forecasts, and (ii) a suite of Python code to visualize the forecasting products. Our products include global visualizations of different atmospheric variables, comparisons of model forecasts with European Center for Medium-Range Weather Forecasts (ECMWF) Reanalysis 5 (ERA5) hourly data \citep{hersbach_era5_single_levels_2023}, and prediction of hurricane tracks. Our work will allow students and instructors to have opportunities to use AI weather forecasting models in university classrooms. This will be an extremely valuable learning experience for students. Specifically, our paper includes examples of 10-day forecasting zonal velocity field at 10 meters vertical height (U10) and a 3-day track for Hurricane Florence from 13 September 2018 to 16 September 2018.

We envision that the materials in this paper will help enable a greater number of researchers, particularly students and junior scientists, to use AI weather forecasting models. To test the effectiveness of our AI model guidance, we trained a group of researchers from City University of New York in February 2025.

Section 2 describes the methods and data used in this paper, Section 3 contains the results of our experimental forecasts using FourCastNetv2, Section 4 introduces a sample of a tutorial session, Section 5 describes a user's manual to make weather forecasts using FourCastNetv2, discussion and conclusions are included in Section 6, and data and code availability in Section 7.

\section{Method and Data}
\subsection{Hardware Setup}
Running FourCastNet on powerful GPUs requires nontrivial setup of relevant hardware. This is a major challenge for leveraging large AI models. This challenge should not be underestimated by aspiring university research groups. The AI weather forecasting groups led by a Principal Investigator should build a work connection with the university's hardware IT team. Figure 1 shows a comprehensive possible interaction among the students and AI research groups with the IT team and AI specialists, such as NVIDIA's FourCastNet developers.

\begin{figure}[h]
    \centering
    \includegraphics[width=0.7\linewidth]{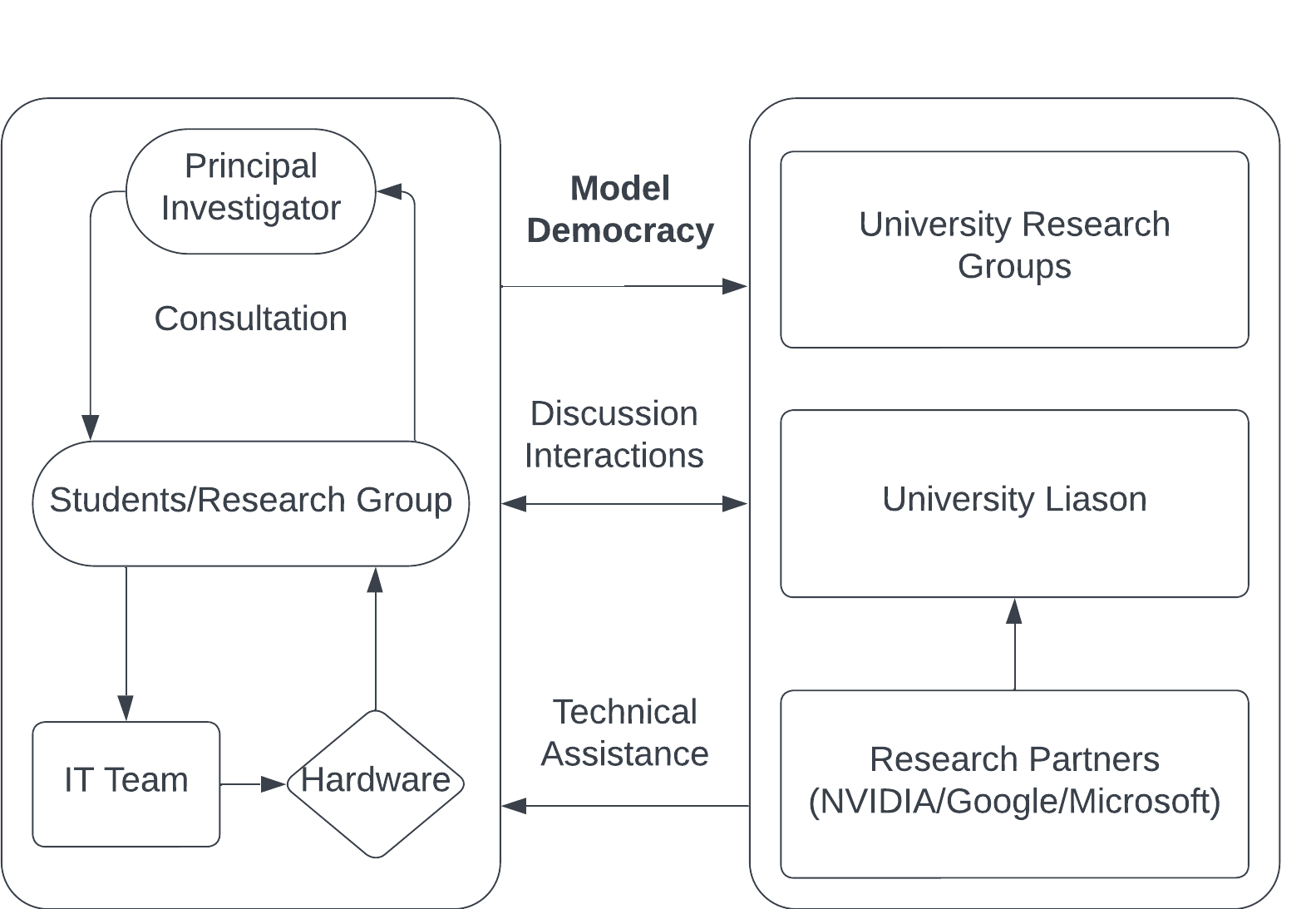}
    \caption{Interactions between groups to support model democracy.}
    \label{fig:enter-label}
\end{figure}

Figure 2 shows the configuration of the hardware for our research. Through the NSF-funded Technology Infrastructure for Data Exploration (TIDE) project, a part of the National Research Platform (NRP) Nautilus hyper-cluster, the San Diego State University (SDSU) Climate Informatics Lab has access up to 4 A100 and 68 L40 NVIDIA GPUs at a time. SDSU’s research IT team has leveraged Kubernetes to optimize the utilization of hardware resources, allowing researchers to easily instantiate and access isolated computing environments within the cluster. Kubernetes is an open-source system for automating the deployment and scaling of applications. Anyone with access can instantiate and gain access to an isolated pod within the cluster. The pods are configured with the packages necessary to interface the hardware with FourCastNet and FourCastNetv2. The usage of isolated pods ensures that each researcher operates in a self-contained environment, preventing interference and resource contention from other users. Moreover, each pod can be configured to use different or multiple GPUs allowing researchers to dynamically scale projects up or down. Jupyterlab was also installed and configured to run inside of each pod so that researchers can quickly manage, edit, and run their code without having to learn Linux terminal commands. The creation of an isolated pod and the configuration of a Jupyterlab allows for new research teams or students to quickly deploy a working environment without the tedious task of chasing down dependencies and, therefore, accelerating the speed of research. 

TIDE provides San Diego State University researchers with free access to research-class computational resources focused on AI and machine learning (ML). TIDE offers 17 GPU nodes each equipped with 4 NVIDIA L40 48 GB VRAM GPUs for a total of 68 L40 GPUs. In addition to the GPU nodes, TIDE has one GPU-Advanced node equipped with 4 NVIDIA A100 80 GB VRAM GPUs. For workloads that do not require access to GPUs, TIDE includes 6 CPU nodes with 64 CPUs each totaling 384 CPUs. The CPU nodes are also equipped with higher amounts of RAM to support CPU-only workloads. TIDE has 3 storage nodes that are configured as a Ceph cluster with triple replication and offers a combined total of 240 TB usable storage. TIDE leverages a 100Gbps HPR connection via Science DMZ to facilitate research network connectivity with other institutions. TIDE also has a connection to the public internet for accessing and publishing public datasets. TIDE is integrated into the National Research Platform (NRP) Nautilus distributed Kubernetes cluster. Kubernetes provides management of storage, RAM, CPU, and GPU requests. Code is packaged into software containers which are then run in Kubernetes via both interactive and batch jobs.

\begin{figure}[h]
    \centering
    \includegraphics[width=0.7\linewidth]{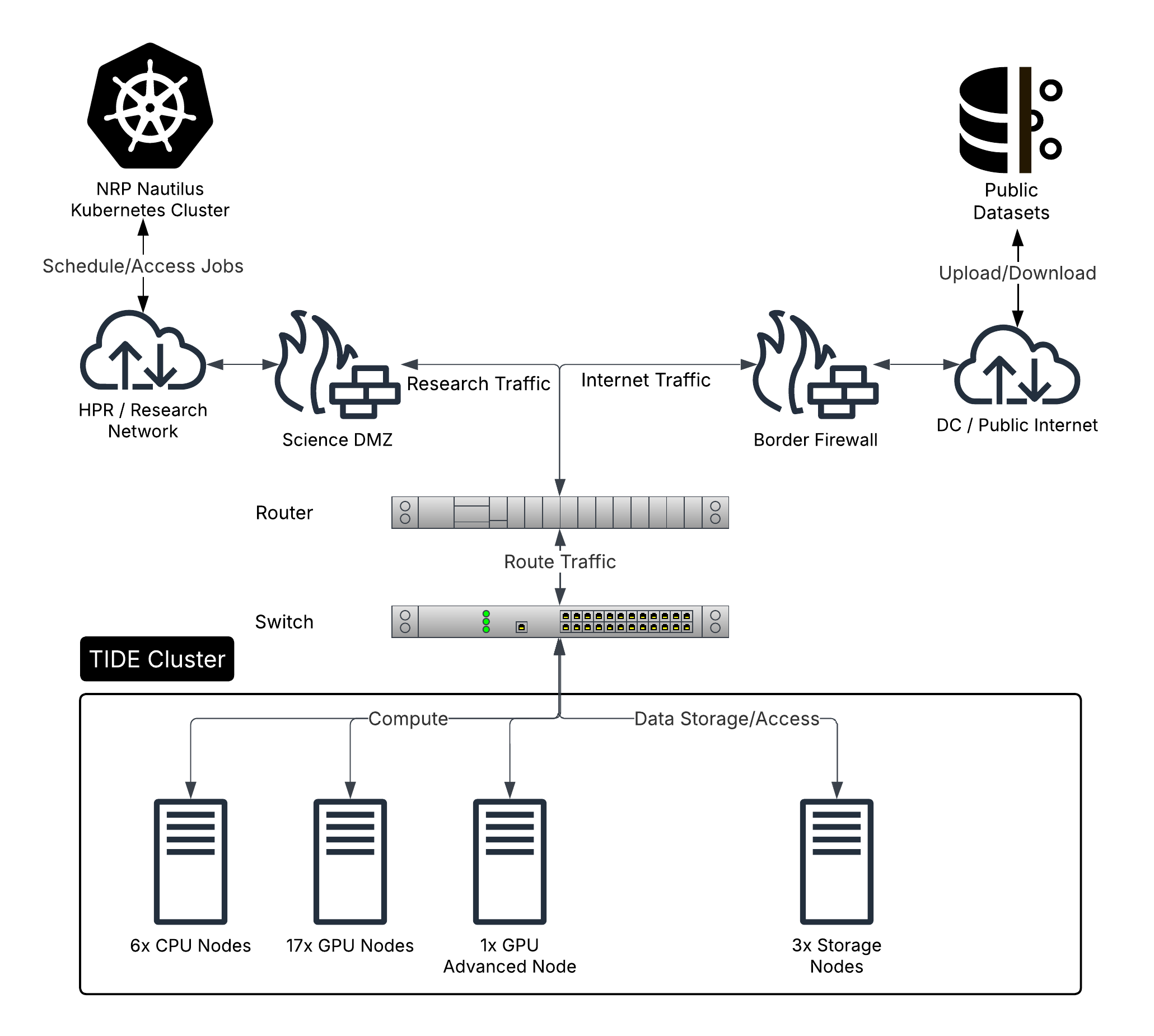}
    \caption{TIDE hardware/software configurations.}
    \label{fig:enter-label}
\end{figure}

\subsection{Training Data}
FourCastNet and FourCastNetv2, like any other AI models, require training data. In this paper we use the ERA5 dataset as the training data, following FourCastNet documentation. ERA5 provides global hourly estimates of weather and climate data with a spatial latitude-longitude resolution of 0.25° $\times$ 0.25°, or approximately 31 km, spanning time from 1940 to present. Covering a broad range of variables such as temperature, wind, pressure, precipitation, and humidity, it offers a comprehensive set of atmospheric and land surface conditions that are invaluable for applications like climate modeling, weather forecasting, and environmental research.

FourCastNet and FourCastNetv2 both utilize the ERA5 dataset for training, but they differ in the number of variables selected for inclusion. Specifically, FourCastNet uses a subset of 20 variables, while FourCastNetv2 leverages a broader subset of 73 variables. Both models use a 6 hour subsample of the hourly ERA5 dataset.

\subsection{Storage of the Training Data}
Approximately 6 terabytes in size, the ERA5 Reanalysis dataset was downloaded from the Copernicus Climate Store by the San Diego State University IT team using Globus. The data is stored in a shared CephFS volume—a distributed, POSIX-compliant file system provided by the Ceph storage platform—inside the TIDE cluster, ensuring accessibility for all project researchers. Each researcher has their own pod with different computation resources, each pod has access to this same shared volume and data. Within such a shared volume are also NVIDIA's model parameters as well as key statistical measures of the data which have been precomputed from ERA5.

\subsection{Initialization of the forecast}

The initialization of forecasts using FourCastNetv2 and FourCastNet begins by accessing a single time step of the required variables in the ERA5 dataset. This can be achieved in many ways, including the Climate Data Store (CDS) API, ECMWF Web Interface, Amazon Web Services (AWS), Google Earth Engine, or simply accessing local files given the dataset is downloaded locally. For this project, we chose to use the CDS API and our local copy of the ERA5 dataset.

To access the CDS API, users must first create an account, install the CDS API, and place the provided authentication key in the designated location on their local device’s file system. Once access to the API is successfully established, users can query it to download the necessary variables. Detailed instructions and code to facilitate this process are available in our user’s manual on Github \citep{khadstev_github}. After the required data has been downloaded, the next step is to arrange the required variables into data arrays with the specific dimensions for the AI forecasting models to process. For FourCastNetv2, the input data array must have the dimension $73\times720\times1440$, while for FourCastNet, the required data array dimension is $20\times720\times1440$. The first dimension corresponds to the set of variables utilized by each model during inference, and the latter two dimensions represent the 0.25° spatial resolution of the ERA5 dataset.

Once the dataset has been gathered and organized into the correct dimensions, the data must be normalized before it is given to the model. This step involves normalizing the input data to ensure that all variables have a mean of $0$ and a standard deviation of $1$. This normalization is performed using the following formula:
\begin{equation}
\hspace{2cm} \text{Normalized Data} = \frac{\text{Raw Data}-\text{Mean}}{\text{Standard Deviation}}
\end{equation}

\vspace{0.3cm}

For each of the variables included in the models, the mean and standard deviations are precomputed across the entire training dataset. This precomputation step is essential as it ensures consistency during both training and prediction, enabling reproducibility and facilitating comparisons between different numerical experiments.

The normalization process, also known as data standardization in meteorology, aligns the scale of each variable, preventing variables with larger numerical ranges, such as wind speed or pressure, from disproportionately influencing the optimization process. This alignment improves the model's ability to learn meaningful patterns from the data and speeds up convergence during training.

After each of the variables has been normalized, FourCastNetv2 requires a different process than FourCastNet to produce the final forecasting products. For FourCastNet, the model needs to be instantiated using the model's source code. Once initialized, users should pass the normalized dataset, the index of the variable they wish to forecast, the desired prediction length, and the model itself into NVIDIA's autoregressive influencer pipeline. This then generates the forecast by iteratively using model output to initialize new predictions. The code required to accomplish this step is documented in detail in our user’s manual (Khadir and Stevenson, 2025).

For FourCastNetv2, users can access the model through the ECMWF Experimental AI-Models API \citep{Raoult_ai-models}, rather than working directly with the model’s source code. This API simplifies the process by automatically handling the normalization of input data, eliminating the need for users to perform this step manually. Additionally, users can integrate the ECMWF API with their CDS credentials, allowing the AI-Models API to handle the data download process as well. A comprehensive guide to setting up and using the API is provided in our user’s manual.

\section{Results}

To show practical AI forecasting scenarios, we chose to present two cases, one is a three-day hurricane track prediction, and another is on 24-hour and 150-hour predictions of two meteorological variables: zonal wind at 10 m height, and surface atmospheric pressure.

\subsection{Predictions of Hurricane Florence}

In the Python Notebook "FourCastNet: A practical introduction to a state-of-the-art deep learning global weather emulator," the original developers of FourCastNet performed a prediction of 2018's Hurricane Michael. To show how students can utilize NVIDIA GPU's and AI technology, we performed two predictions of 2018's Hurricane Florence, one by FourCastNet, and another by FourCastNetv2.

\subsubsection{Model initialization and validation data}

\noindent\textit{(a) Data for the initial conditions} 

Forecast initialization with FourCastNet and FourCastNetv2 requires a global one-hour mean dataset comprising all 20 variables for FourCastNet and 73 variables for FourCastNetv2. All input variables are needed when to perform a forecast. By default, the forecast is for every variable. Namely, when running FourCastNetv2, all 73 variables are being predicted. A user may choose to output the forecast result for one or more variables, such as mean sea level pressure and 10-meter zonal wind. Table 1 lists the 73 required variables of FourCastNetv2. 

\begin{table}[h!]
\centering
\textbf{Table1.} ~ FourCastNetv2's 73 variables.
\vspace{0.2cm}

\begin{tabular}{|c|l|c|}
\hline
\textbf{Variable} & \textbf{Description} & \textbf{Pressure level (hPa)} \\
\hline
u10 & 10 meter u-wind component & -- \\
u100 & 100 meter u-wind component & -- \\
v10 & 10 meter v-wind component & -- \\
v100 & 100 meter v-wind component & -- \\
t2 & 2 meter temperature & -- \\
sp & Surface pressure & -- \\
msl & Mean sea level pressure & -- \\
tcwv & Total column vertically-integrated water vapor & -- \\
\hline
z--- & Geopotential (at pressure level ---) & \multirow{5}{*}{\parbox{4.5cm}{50, 100, 150, 200, 250, 300,\\400, 500, 600, 700,\\850, 925, 1000}} \\
t--- & Temperature (at pressure level ---) & \\
u--- & u component of the wind (at pressure level ---) & \\
v--- & v component of the wind (at pressure level ---) & \\
r--- & Relative humidity (at pressure level ---) & \\
\hline
\end{tabular}\\
\hspace{1cm}
\label{tab:ecmwf_variables}
\end{table}

Table 1's first eight variables, such as 10-meter zonal u--wind component and mean sea level pressure (SLP),
 in the upper half of the table are variables on a single layer of the atmosphere. The five parameters, such as temperature and 
 relative humidity, are for 13 layers of the atmosphere, divided according to the atmospheric pressure 
 levels, 50, 100, $\cdots$, 1,000 hPa. These five parameters at the 13 layers are counted as 65 variables ($65 = 5\times 13$).
 Thus, altogether we have 73, which is 8 plus 65, variables for FourCastNetv2.
 
\noindent\textit{(b) Validation design} 

We chose to use Hurricane Florence as an example in this paper. Hurricane Florence occurred in 2018, which is the year designated for validation and therefore excluded from the training sets of both FourCastNet and FourCastNetv2. This ensures that the models have not previously “seen” the event during training, allowing for an objective evaluation of their forecasting accuracy. This is the same reason why 2018's Hurricane Michael was originally selected by Pathak et al. (2022) for their forecast validation.

Our validation will be focused on the hurricane track. The predictants are mean sea level pressures (SLP) for a 102-hour period starting from September 13th, 2018, at 12:00 AM, with predictions made in 6-hour intervals. After generating the predictions, we identify the eye of the hurricane by locating the minimum mean SLP at each 6-hour interval.

\subsubsection{Prediction Results}

Figure 3 presents two different predictions of hurricane dynamics from two different models: 
Fig. 3 (a) for FourCastNet,  and Fig. 3(b) FourCastNetv2. The initial conditions are shown by the background color map for mean SLP and by the black arrows for the wind field at 12:00 AM, September 13th, 2018. The mean SLP field at 12:00 AM, September 13th, 2018 is the initial condition for one of the 73 variables. We use the gridded $0.25^\circ \times 0.25^\circ$ lat-lon hourly ECMWF ERA5 data \citep{hersbach_era5_single_levels_2023} for all 73 variables at 12:00 AM, September 13th, 2018, as the complete initial condition to make our FourCastNetv2 predictions.

\begin{figure}[H]
        \centering
        \includegraphics[width=0.9\linewidth]{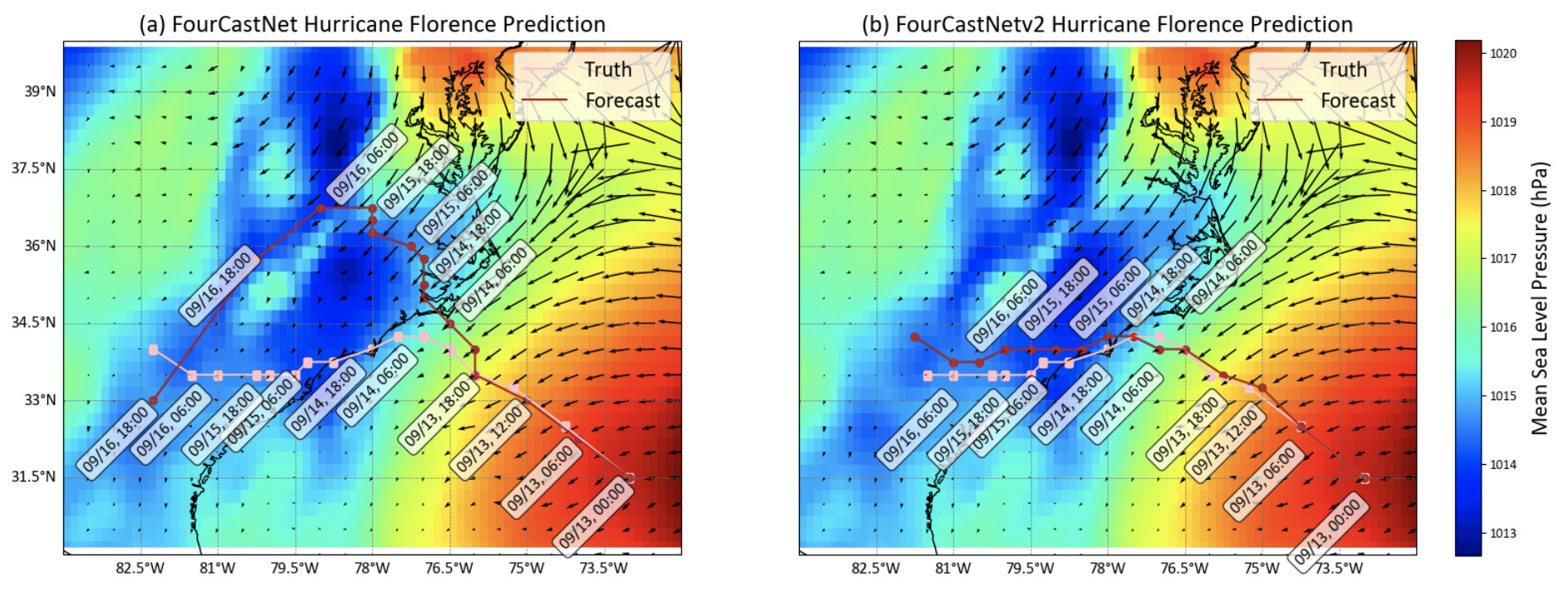}
        \caption{(a) FourCastNet Hurricane Florence Prediction and (b) FourCastNetv2 Hurricane Florence Prediction.}
\end{figure}

The predictions are the locations of the hurricane eyes, which are defined as the locations of minimum sea level pressure at a given time and represented by the red dots for predictions. 
The predicted hurricane track, for approximately four days from 12:00 AM, September 13 to 18:00, September 16,
  is compared with the ground truth represented by the pink dots. See Fig. 3.

Figure 3 (a) shows a significant deviation of the FourCastNet's forecasts from the truth after the hurricane made landfall. The FourCastNet model fails to maintain an accurate representation of the hurricane's trajectory over land, which may be attributed to limitations in its ability to handle changes in environmental conditions that occur during landfall.

In contrast, FourCastNetV2 exhibits superior performance throughout the entire lifecycle of the hurricane (see Fig. 3 (b)). The prediction remains accurate even after landfall, suggesting that FourCastNetV2 is better equipped to handle the complexities of land interaction. This improvement is explained by advancements in model architecture, the advent of the Spherical Fourier Neural Operator \citep{bonev2023sphericalfourierneuraloperators} and the more robust training FourCastNetv2 underwent compared to FourCastNet.

Following the tutorial guidance we have provided in the GitHub repository \citep{khadstev_github}, a student can make the same predictions and reproduce Fig. 3. This demonstrates what undergraduate students and low resource research groups can achieve when given guided access to powerful AI tools, such as FourCastNet and FourCastNetv2. Here, the computer code for data visualization and the model run were entirely made by an undergraduate student research with references to the existing climate data visualization code in modern textbooks and their online materials, e.g., \citet{shen2023statistics} and \citet{shen2019climate}. The visualization code and guidance to access the correct data is included in our user's manual in GitHub \citep{khadstev_github}.

\subsection{Global Predictions}

\subsubsection{Zonal Wind Predictions}
Using ECMWF's "ai-models" API, we leverage FourCastNetv2 to forecast zonal winds and surface pressure.Figure 4 presents two predictions of global zonal winds from FourCastNetv2 at 24 hours and 150 hours.


\begin{figure}[h!]
    \centering
    \includegraphics[width=0.9\linewidth]{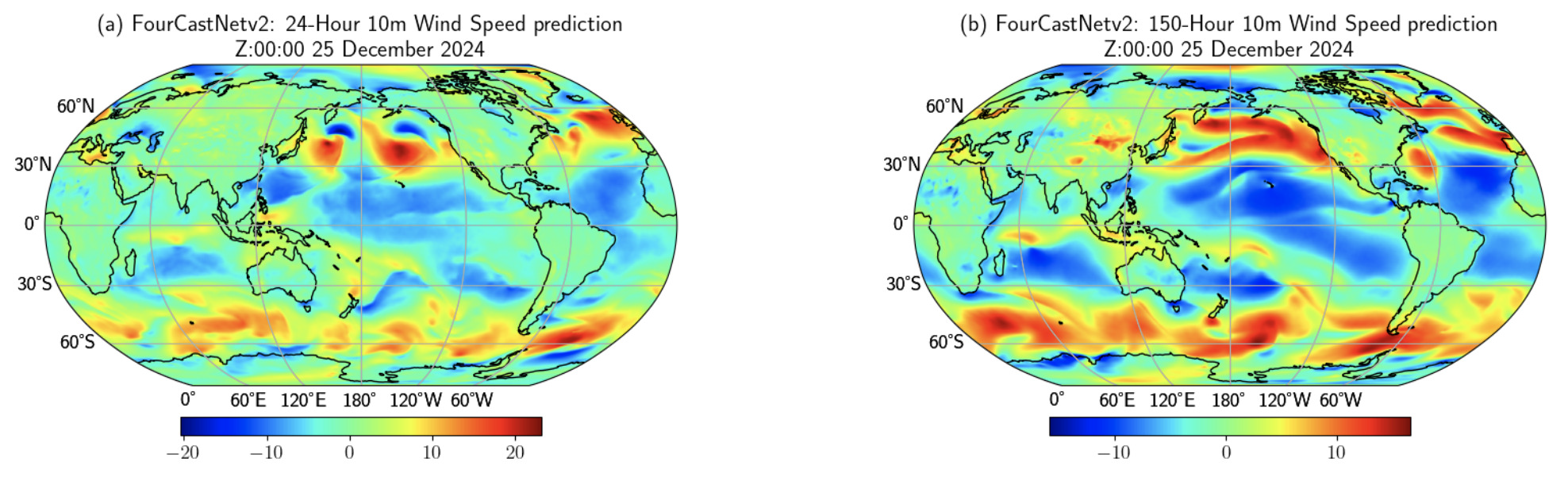}
    \caption{Two FourCastNetv2 predictions of zonal winds u10 from the initial time Z00:00 25 December 2024: (a) 24-hour prediction, (b) 150-hour prediction.}
    \label{fig:winds}
\end{figure}

The 24-hour zonal wind predictions demonstrate a high level of detail, reflecting the model's strong performance with shorter forecasts. Coherent and sharp wind patterns indicate the model's ability to capture and simulate the fine-scale structures of the atmosphere. For example, the zonal wind patterns associated with the Intertropical Convergence Zone (ITCZ) and the South Pacific Convergence Zone (SPCZ) are well maintained. The mid-latitude westerlies over both the Pacific and Atlantic are clearly persistent. Even after 150 hours both patterns are still clearly present. The preservation of key global wind patterns highlights the model's robustness over an extended time frame.

\subsubsection{Surface Pressure Predictions}

Figure 5 presents two predictions of surface pressure from FourCastNetv2 at 24 hours and 150 hours from the initial time Z00:00 25 December 2024.

\begin{figure}[H]
    \centering
    \includegraphics[width=0.9\linewidth]{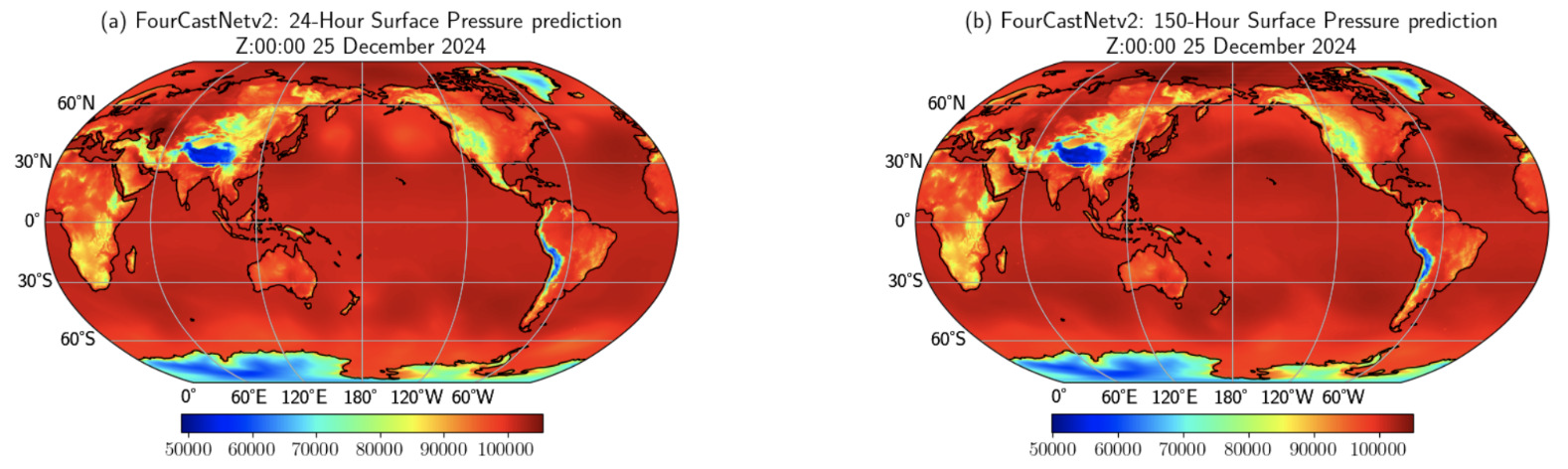}
    \caption{Comparison of surface pressure from initial time of Z00:00 25 December 2024: (a) 24-hour prediction, (b) 150-hour prediction.}
    \label{fig:sp}
\end{figure}

The model clearly represents the characteristically low surface atmospheric pressure over the Tibetan Plateau, the Andes, and the North American Rockies in both the 24-hour and 150-hour predictions.
The atmospheric pressure over the sea surface and relatively lower elevation land areas are maintained around 1,000 hPa.

\subsubsection{Accessibility and Practical Impact of FourCastNetV2}

These results highlight the model's utility for long-range forecasting, an area where maintaining physical coherence is challenging due to the chaotic nature of the atmosphere. Furthermore, the consistency across time ranges illustrates the potential for using advanced AI-based forecasting systems in operational and research contexts.

Furthermore, these forecasts are achievable by a typical computer owned by a college student, so long as the user has access to the Climate Data Store (CDS). FourCastNetV2 is designed to be easily accessible, requiring only modest computational resources to run. This helps democratize high resolution global weather forecasting, allowing researchers, educators, and even hobbyists to make high-quality global predictions without the need for expensive supercomputers.

The ease of use and low barriers to entry open up numerous possibilities. For example, undergraduate students can leverage FourCastNetV2 for projects and research, as demonstrated in this paper. Additionally, it enables low-resource institutions or organizations in developing countries and regions to improve their understanding of local weather patterns and better prepare for extreme events.

The code to achieve the depicted visualizations, along with detailed instructions, is included in our user manual, ensuring that users can not only replicate our results but also produce their desired forecasts. This accessibility provides opportunities for making numerical weather predictions in classrooms.

\subsection{Setup and Results for Training FourCastNet}

Thus far, we have primarily explored the power of inference, demonstrating how FourCastNet and FourCastNetv2 can be leveraged by undergraduates and resource-limited research groups for meaningful contributions to atmospheric research. Another important aspect of utilizing these AI models is training. In this paper, we use FourCastNet as an example to describe 
 the setup and results of training.

Model training requires substantial computational resources, particularly for large-scale AI-driven weather forecasting. The original FourCastNet model was trained on a cluster of 64 A100 GPUs over 16 hours \citep{pathak2022fourcastnetglobaldatadrivenhighresolution}, a setup far beyond the reach of many research institutions. However, advancements in computing accessibility have enabled universities and organizations to allocate powerful hardware to their students. San Diego State University's NRP Hyper Cluster is one such example, providing access to high-performance GPUs for training deep-learning models.

Using four A100 GPUs from the NRP Hyper Cluster, we fine-tuned FourCastNet’s backbone for an additional 18 epochs, spanning 110 hours of training with a batch size of 16, significantly smaller than NVIDIA’s original batch size of 64. To adapt the model to our system, we modified several scripts and configurations from the original FourCastNet release. The model was trained on ERA-5 global data from 1979 to 2015, with the years 2016 and 2017 reserved for testing, and 2018 held out as a dedicated validation set to assess its generalization performance on unseen data. Figure 6 shows validation loss across the 18 training epochs when training FourCastNet. 
A complete guide to reproducing our training setup, including the necessary configurations and environment details, is available in our user manual on GitHub \citep{khadstev_github}.

\begin{figure}[H]
    \centering
    \includegraphics[width=0.6\linewidth]{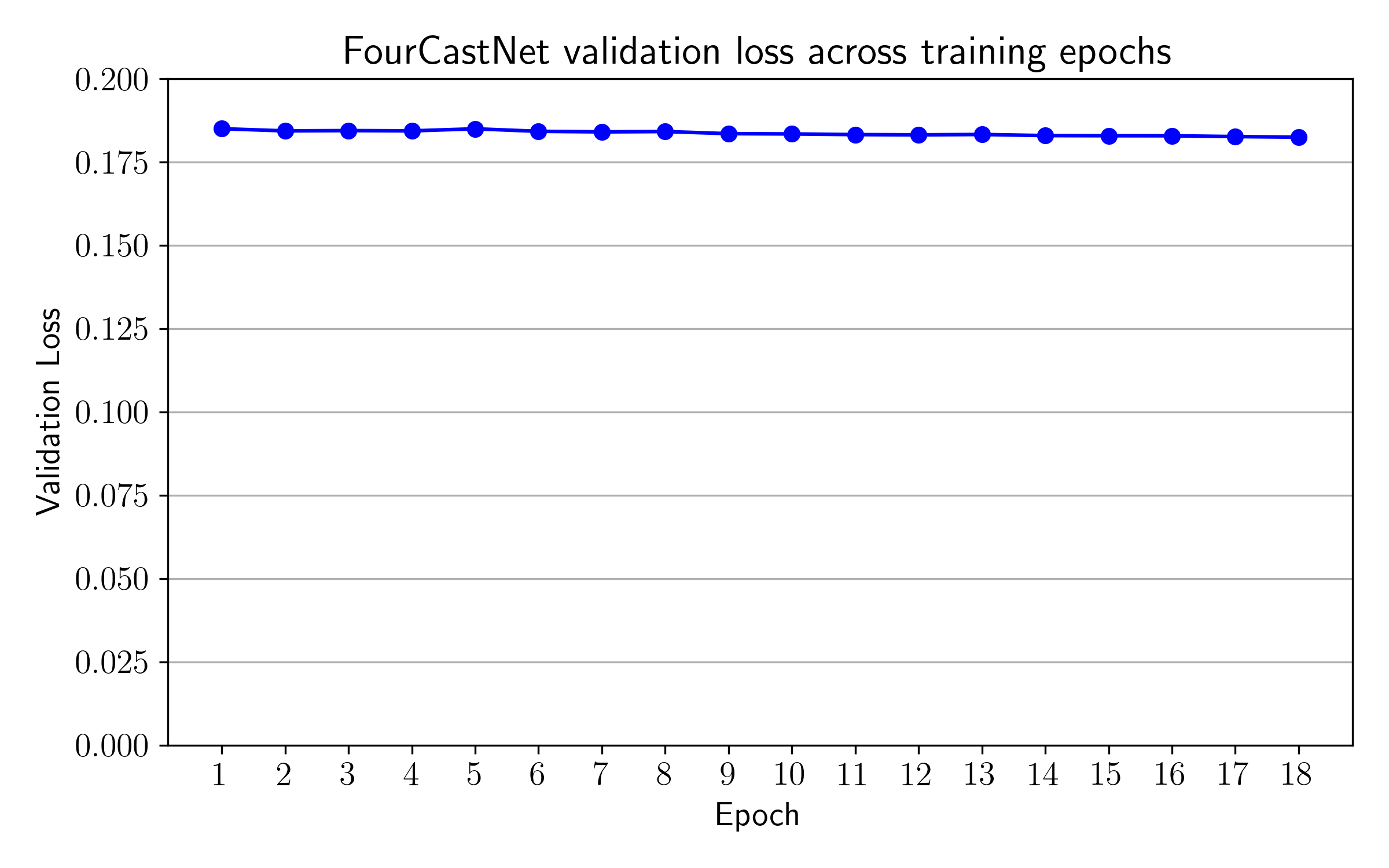}
    \caption{FourCastNet validation loss across 18 training epochs.}
    \label{fig:enter-label}
\end{figure}

In "FourCastNeXt: Optimizing FourCastNet Training for Limited Compute" \citep{guo2024fourcastnextoptimizingfourcastnettraining}, the authors found that improved results were only achieved when training on a more diverse dataset, allowing FourCastNet to continue refining its predictions. Our experience confirmed this, as throughout all 18 epochs, both the training and validation loss oscillated around the same value. We theorize that this plateau was not due to insufficient training effort, but because the model had already converged on the provided data and weights. Figure 6 illustrates the validation loss for FourCastNet across these 18 epochs, where one epoch represents a pass through the entire dataset (ERA-5 1979 to 2015). The points show the mean squared error (MSE) between FourCastNet's predictions and the ground truth of the validation dataset. The stagnation of the model's validation loss indicates that it was unable to improve on the current dataset.

The ability to connect our A100 GPUs to FourCastNet's complex system for AI model training is another significant step. A major barrier for undergraduate students and resource-limited research groups is the technical knowledge required to set up and effectively utilize advanced GPU systems. Integrating one system with another to enable the use of AI models for training is often the most challenging task in the entire AI weather forecasting process. For example, the FourCastNet scripts were designed for a Simple Linux Utility for Resource Management (SLURM) based system, while the system we had access to was Kubernetes based. This mismatch required a deep understanding of both systems and the ability to translate the necessary configurations and scripts from one environment to another. The technical expertise to adapt these resources and integrate them seamlessly is a key capability for researchers who wish to leverage AI for atmospheric research. The close collaboration of San Diego State University's Research and Cyberinfrastructure team allowed for us to interact with the necessary resources through a Kubernetes environment specially created with the appropriate packages installed to use FourCastNet (see Figs. 1 and 2 about the setup of the collaboration and hardware). By successfully configuring and adapting FourCastNet to work with our available hardware, we have demonstrated that even researchers with limited expertise in GPU systems can still leverage high-performance resources for deep-learning tasks. Ultimately our goal is to lower the entry barriers and empower a broader community of researchers to explore and utilize AI-driven models, which is crucial for advancing AI in fields like weather forecasting.

\section{A Sample Tutorial Session for FourCastNetv2}

To illustrate the accessibility of NVIDIA's FourCastNetv2, this section describes our experience of running a hands-on tutorial session, which was conducted between the San Diego State University (SDSU) undergraduate authors (Shane Stevenson and Iman Khadir) of this paper and researchers at both SDSU and the City College of New York (CCNY). The goal of the session was to equip participants with the skills necessary to leverage FourCastNetv2 for weather forecasting and climate research. Facilitated by Professors Samuel Shen of SDSU and Mitchell Goldberg of CCNY, the session brought together students and faculty to learn how to use FourCastNetv2. Participants had access to a range of hardware, from personal computers to the CUNY High Performance Computing Cluster (HPCC). Some CUNY participants had access to the powerful HPCC computing resources, including 8 H100 80GB VRAM GPUs, 2TB of RAM, and 30TB of system storage, as well as extremely high-speed internet. Other participants followed along on their consumer-grade personal computers. 

The trainers (Shane Stevenson and Iman Khadir) shared their presentation slides via GitHub\\

\textit{https://github.com/ikhadir/FourCastNet-with-JupyterLabs/blob/main/Leveraging\_FourCastNetv2.pdf}\\

The trainees could access the slides and copy-and-paste the Internet links used in the training session. 

The first step in the tutorial was to install the necessary packages and frameworks to interface with FourCastNetv2. This included Python, PIP, a Climate Data Store (CDS) account, and ECMWF's "ai-models" Python package.

The second step involved accessing the data required to run FourCastNetv2, including NVIDIA's precomputed global means and global standard deviations across the entire ERA5 time history. This was easily done through a simple curl command in the participants' command line interface, e.g., on the Terminal in Mac. 

However, downloading the model weights presented a challenge. While this was also done through a curl command, the weights are over 3 GB in size, meaning participants with slower internet speeds struggled to keep up. Those with access to the CUNY HPCC were able to quickly download FourCastNetv2's model weights without issue. For future sessions, we plan to ask participants with consumer-grade computers to download the model weights ahead of time to avoid this bottleneck.

The third step was to use CDS API Python package. Before performing forecasts with FourCastNetv2, participants needed to pull data from CDS. This was achieved through the CDS API Python package. Participants were shown how to install this package, configure it, and accept the ERA5 data license. Once this was done, they could use ECMWF's "ai-models" package to download the necessary initialization data for FourCastNetv2. With everything in place, participants generated forecasts using FourCastNetv2. If the previous steps were executed correctly, running the forecast was as simple as entering a terminal command specifying FourCastNetv2 and the desired initialization date. The "ai-models" package then downloaded the data from CDS and initialized a 10-day forecast from the specified start date. One CUNY trainee was able to immediately generate a forecast right in the 90-minute training session (see Fig. 7), while some encountered a compatibility issue with their version of PyTorch. After resolving this and updating the information on the tutorial slides, other trainees were able to proceed successfully.
\begin{figure}[h!]
    \centering
    \includegraphics[width=0.55\linewidth]{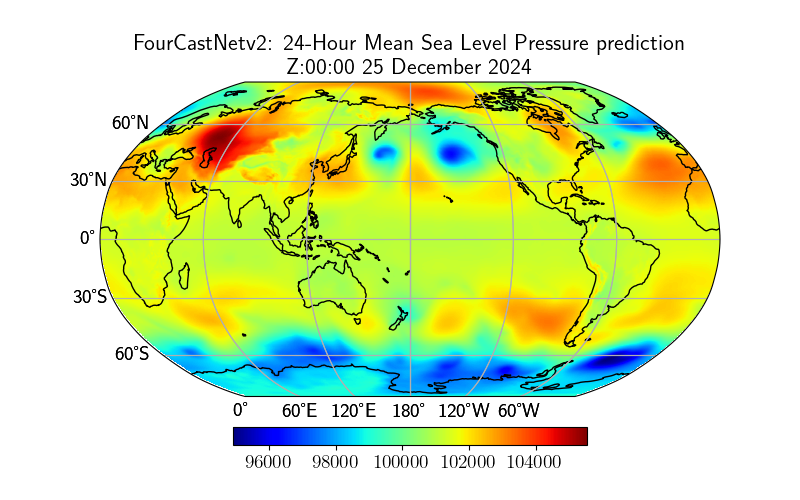}
    \caption{A 24-Hour mean sea level pressure forecast produced by a trainee during the CCNY tutorial session.}
    \label{fig:enter-label}
\end{figure}

After the forecasts were generated, participants were instructed to download several packages for data manipulation and visualization, including NumPy, Xarray, Matplotlib, and Cartopy. They were given Python scripts to plot their forecasts and shown how each script functioned. By the end of the session, the two trainees with access to the CUNY HPCC had completed a full 10-day forecast using FourCastNetv2, and they were instructed on how to visualize and interpret the results. Participants with consumer-grade computers who were unable to download the model weights in time later successfully followed the rest of the tutorial and were able to leverage FourCastNetv2 for weather forecasting.

\section{User's Manual to Make Weather Forecasts Using FourCastNetv2}

We have created a user's manual to further help democratize AI weather forecasting models and guide students to make their own forecasts. The user's manual is available on GitHub at \textit{https://github.com/ikhadir/FourCastNet-with-JupyterLabs}. See Fig. 8 for a screenshot of our tutorial material. 

\begin{figure}[!htb]
    \centering
    \includegraphics[width=1.0\linewidth]{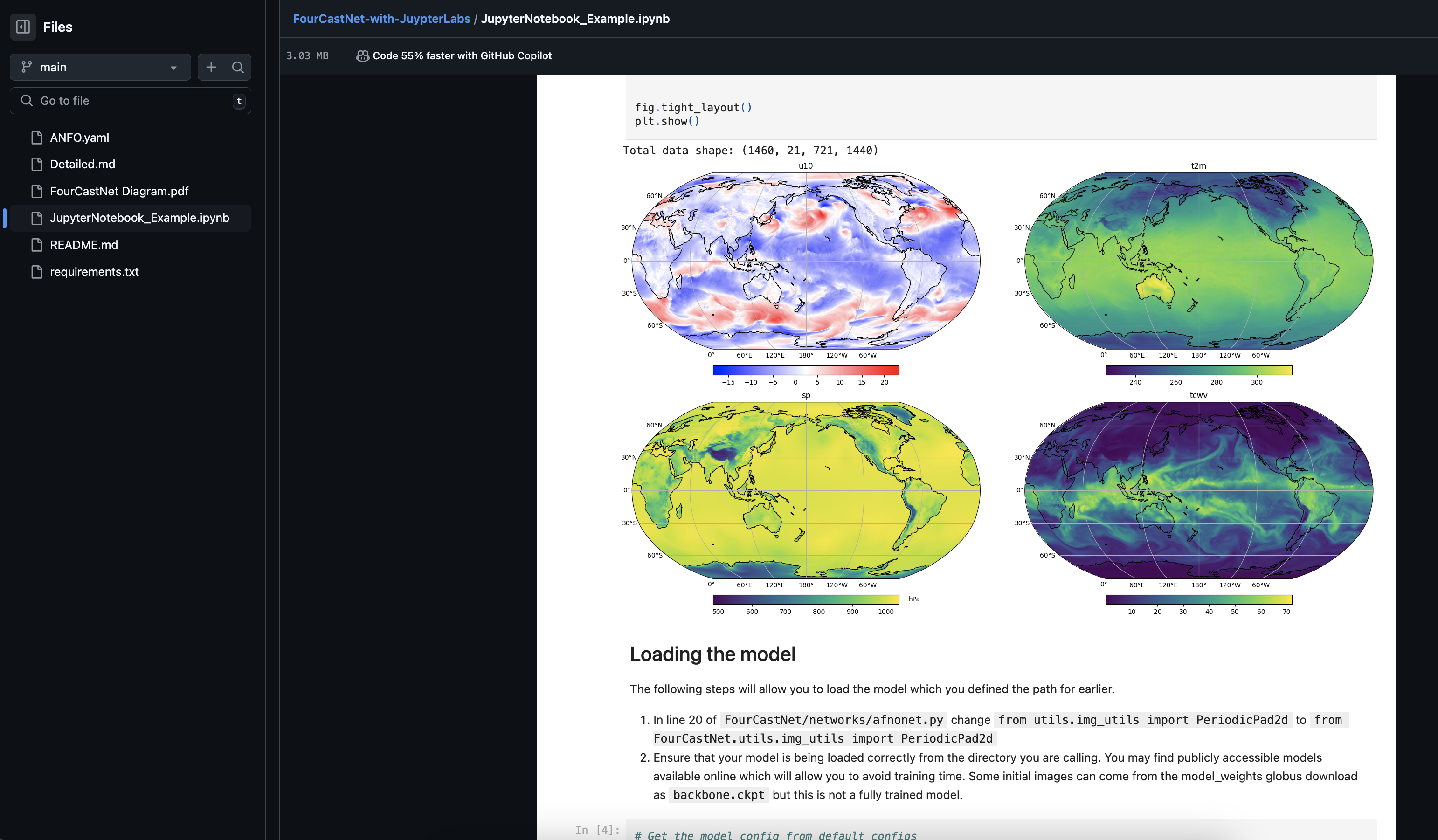}
    \caption{Screenshot from the Jupyter Notebook from the user's manual in the GitHub.}
    \label{fig:enter-label}
\end{figure}

The intention of the user's manual is to allow students to become familiar with the setup and initial implementation of the model. To this end, a Jupyter Notebook has been created to assist students from a variety of backgrounds to pull ERA5 climate data, choose parameters, select date ranges, visualize the data, predict parameters for dates outside of the training data, and visualize and analyze predictions. Students will find this useful because they will not be hindered by long exploration pages of the original NVIDIA's FourCastNet documentation and will be able to jump into the procedures of prediction and visualization immediately. This ease of access to predictions will allow students to explore further the various features of FourCastNetv2. This manual will also serve to increase student training in visualizing large climate data and to garner interest in AI weather models. 

Users of our GitHub repository \textit{https://github.com/ikhadir/FourCastNet-with-JupyterLabs} should make adjustments to our sample code where necessary. Options for visualizations of the data and predictions are available to suit the prediction being created. We visualize the data using a Pacific-centered Robinson projections with grid lines marking longitude and latitude. The color map that is selected can be determined for the specific variable to be shown. We encourage users to tread the manual as a tool to explore the AI forecasting model in its entirety. We have also created tools to improve the visualizations of the raw data, global predictions, and hurricane tracking. These tools are straightforward to use and allow for a multitude of configurations and adjustments to fit the user's preferences and demands.

\section{Discussion and Conclusions}

We find it encouraging that FourCastNetv2 works efficiently  and is accessible to small groups. It has become evident that FourCastNetv2 is an effective tool for university research groups to produce AI weather predictions in a fast and convenient way. FourCastNet is also a subset of the services and software for weather prediction provided by Earth2, an additional NVIDIA product that researchers and students can utilize for AI weather predictions. 

The existing fast and convenient AI forecasting work will likely lead to a long term progress  to democratize AI models and hence give students hands on experience in AI weather predictions. Among numerous opportunities, the following three areas may present immediate and  practical applications for future research exploration: (a) AI weather predictions for small island nations, (b) training students in classrooms using FourCastNet, and (c) generating predictions with real-time satellite data. 

For area (a), we can envision applications of the AI weather forecasting model around the world to solve a variety of problems. One such application is the predictive power of the model for remote islands.  The experience of North Carolina State University meteorology students provided crucial wind forecasts for a team of ecologists on Bouvet Island \citep{2003BAMS...84..777R}. Communication and data traffic to the island were limited due to the bandwidth constraint. Weather predictions were crucial for the completion of the research as well as the safety of the crew, one of whom was already injured by strong winds. Riordan describes the wind predictions his team provided as having marginal reliability at 48-hour lead time. Running FourCastNetv2 locally will 
avoid the problem of slow data transmission and allow island residents to create their own forecasts and act accordingly. 
These forecasts will be valuable to researchers and local populations in preparing for agricultural needs, preventing flash flood damage, and taking appropriate measures for public safety during hazardous weather events.
As an example of this, we provide the U10 predictions for the Pacific Island nation of Vanuatu, which is particularly vulnerable to tropical cyclones (see Fig. 9).

\begin{figure}[h]
    \centering
    \includegraphics[width=1.0\linewidth]{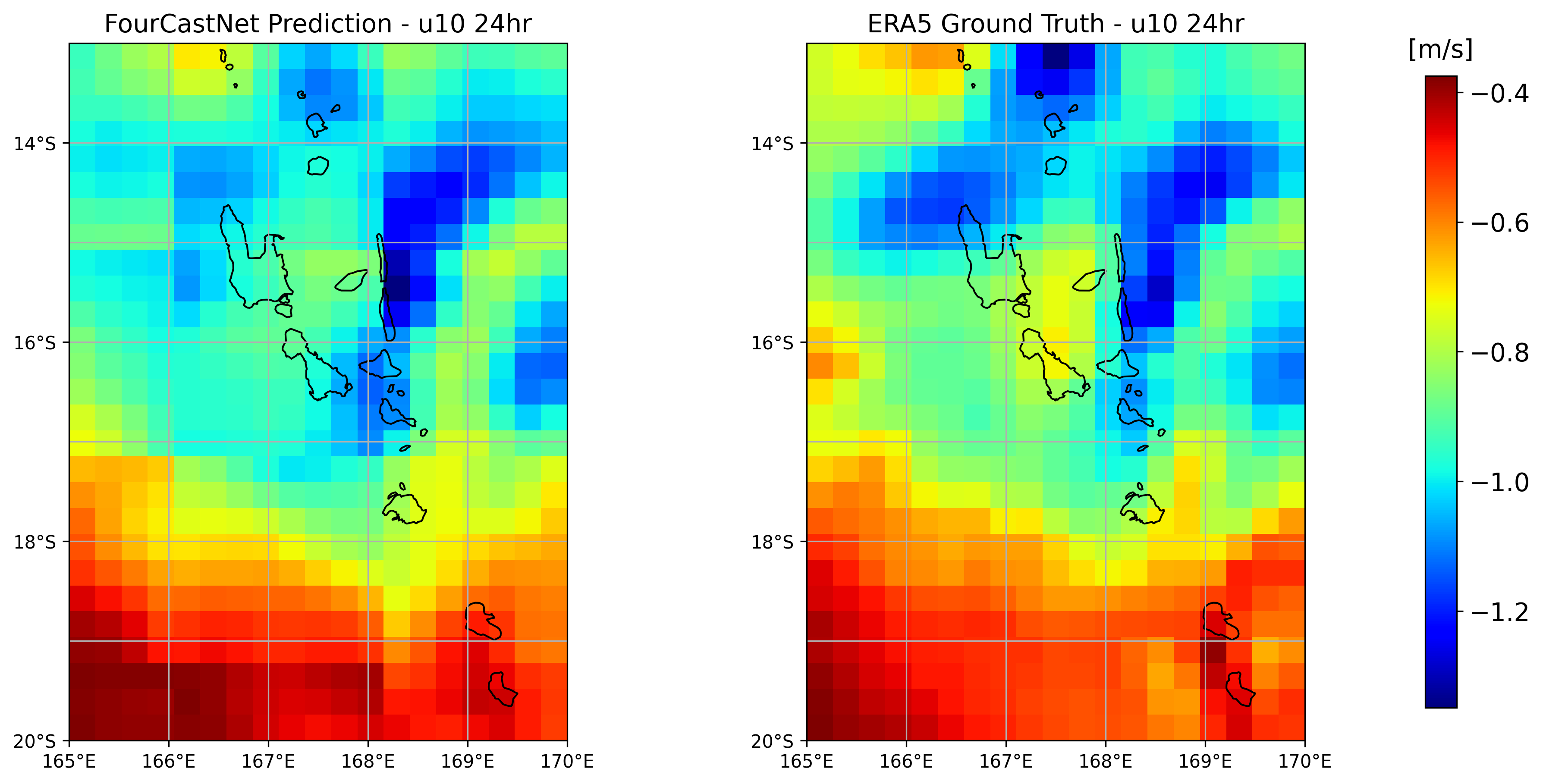}
    \caption{U10 field (units: $ms^{-1}$) for the Pacific island nation of Vanuatu.}
    \label{fig:enter-label}
\end{figure}

For area (b), in this paper, we wish to share our student experience with the FourCastNet AI weather models.  The audience of this paper will find that they benefit better from our experience if they are more familiar with some of the technical details of FourCastNetv2, such as the variables which the model uses and the output file which the predictions produce. The first two authors of this paper are undergraduate students and encountered certain challenges that could be avoided by other university students who implement the model. For a successful implementation, students will need some familiarity with Bash to interact with the local development environment and coordinate hardware configurations with the model itself. As indicated in Figure 1, strong support from a dedicated IT team in a university is crucial for setting up the necessary software and hardware configurations. The SDSU Climate Informatics Lab (SCIL) has been instrumental in providing consultation for this project. It is essential that students have the backing of a supportive principal investigator, such as the SCIL director Samuel Shen in our case. It is also important to have a committed team of lab members who collaboratively ensure that the AI weather forecasting research meets high-quality and professional standards.

The opportunity for using FourCastNet and FourCastNetv2 as a training tool for students to gain experience in AI weather forecasting and machine learning is wide. Many students are often left without the resources or tools to be able to use such large climate datasets in a manner that is exciting and comprehensive. Our work will allow many students to gain valuable training. We have developed supplementary notebooks for students to use ERA5 data to create visualizations. The training notebooks have been expanded from NVIDIA's notebooks to include further visualizations. See our GitHub repository \textit{https://github.com/ikhadir/FourCastNet-with-JupyterLabs}

For area (c),  some university groups may aggregate  
satellite data directly without a data assimilation step and 
use the aggregated data on a 4D space-time grid to allow the initiation of the AI weather forecasting model to perform near real-time forecasts. Thus far, our AI forecastings are limited to use the assimilated ERA5 output as initialization. Although ERA5 data can be pulled using API, there is a five day delay for the present data to be uploaded. Using the space-time gridded satellite data directly will allow for near real time forecasts to be produced.

Our analysis of FourCastNet and FourCastNetv2 demonstrates the potential of advanced AI models for weather prediction, particularly in the context of limited computational resources faced by university research groups. NVIDIA's FourCastNet models drastically reduce the time and cost of creating weather forecasts compared to traditional Numerical Weather Prediction (NWP) methods, but there are still few barriers for small research groups to replicate AI weather forecasting results. Our work intends to resolve some the practical challenges  of using GPUs, such as NVIDIA A100, for large-scale machine learning applications in numerical weather forecasting.

\section{Code and Data Availability}

The initial version of FourCastNet is available under the NVIDIA Source Code License at 
https://github.com/NVlabs/FourCastNet under DOI: https://doi.org/10.5281/zenodo.16389569. FourCastNetv2 is the second version, accessible through the ECMWF AI Models 
Plugin at https://github.com/ecmwf-lab/ai-models under  DOI: https://doi.org/10.5281/zenodo.16389730, released under the Apache License 2.0. 
A tutorial for using the initial version of FourCastNet and FourCastNetv2 is freely
 available at https://github.com/ikhadir/FourCastNet-withJupyterLabs under DOI: https://doi.org/10.5281/zenodo.16118616. The computer code for the tutorial used in this paper is also available at this DOI. We used the following NVIDIA materials: (i) The trained model weights, (ii) the preprocessing scripts, (iii) the ERA5-derived input data. These materials are publicly available via Globus at \url{https://app.globus.org/file-manager?origin_id=945b3c9e-0f8c-11ed-8daf-9f359c660fbd&origin_path=%2F~%2Fdata%2F}. We used these materials to produce the
   results in this paper. The ERA5 reanalysis dataset used for the FourCastNetv2 model training and validation is available from the Copernicus Climate Data Store (CDS) at https://cds.climate.copernicus.eu upon account registration.
     We downloaded approximately 6TB of input data from NVIDIA's Globus folder for our study in this paper.

\section{Author contribution}

IK, SS, and AB developed the Jupyter notebook tutorial; HL and KK created the relevant cyberinfrastructure to connect to SDSU GPU resources; IK, SS, and SSPS wrote the manuscript draft; HL, KK, DH, and SP reviewed and edited the manuscript. All coauthors have contributed to the original ideas and research objectives equally.

\section{Competing interests}

The authors declare that they have no conflict of interest.

\section{Acknowledgments}
We thank the collaboration and support from the colleagues at SDSU Climate Informatics Laboratory, in particular, the indispensable guidance of Danielle Lafarga. The Research and Cyberinfrastructure group at SDSU provided constant technical support for hardware setup and software assimilation. We thank researchers at ECMWF for maintaining the ERA5 dataset. Sincere thanks to Jacob Radford from the University of Colorado for his guidance in running the ai-models plug-in. We thank Mitch Goldberg's group at City University of New York for the opportunity to demonstrate our work. Finally, we thank Amy McGovern, Director the U.S. National Science Foundation AI2ES project (Award No. 2019758), for bridging the collaboration between the SDSU and NVIDIA researchers on this work. 

\section{Financial Support}

This research used resources from the U.S. National Science Foundation (NSF)-funded Technology Infrastructure for Data Exploration (TIDE) project (Award No. 2346701) and from the the National Research Platform (NRP) at the University of California, San Diego (NSF award No. 1730158, 1540112, 1541349, 1826967, 2112167, 2100237, and 2120019). This work was also partially
supported by a U.S. NSF grant (Award No. IIS 2324008) and  U.S. NOAA grants (Award No. NA22SEC4810016 and NA24NESX405C0006-T1-01).

\bibliography{references}  





\end{document}